\documentclass{article} % For LaTeX2e
\usepackage{nips14submit_e,times,graphicx}
\usepackage{url}
\usepackage{amsmath}
\usepackage{amssymb}
\usepackage{booktabs,dcolumn}
\usepackage{microtype}
\newcolumntype{d}[1]{D{.}{.}{-#1}}

\title{Semi-supervised Sequence Learning}

\author{Andrew M. Dai \\
Google Inc. \\
\texttt{adai@google.com} \\
\And
Quoc V. Le \\
Google Inc. \\
\texttt{qvl@google.com}
}

\nipsfinalcopy % Uncomment for camera-ready version

\begin{document}

\maketitle

\begin{abstract}
We present two approaches that use unlabeled data to improve sequence
learning with recurrent networks. The first approach is to predict
what comes next in a sequence, which is a conventional language model
in natural language processing. The second approach is to use a sequence
autoencoder, which reads the input sequence into a vector and predicts
the input sequence again. These two algorithms can be used as a
``pretraining'' step for a later supervised sequence learning
algorithm. In other words, the parameters obtained from the
unsupervised step can be used as a starting point for other supervised
training models. In our experiments, we find that long short term
memory recurrent networks after being pretrained with the two
approaches are more stable and generalize better. With pretraining, we
are able to train long short term memory recurrent networks up to a few hundred
timesteps, thereby achieving strong performance in many text
classification tasks, such as IMDB, DBpedia and 20 Newsgroups.
\end{abstract}

\section{Introduction}
Recurrent neural networks (RNNs) are powerful tools for modeling
sequential data, yet training them by back-propagation through
time~\cite{werbos1974beyond,rumelhart1986learning}
can be difficult~\cite{hoch01}. 
%Many approaches have been proposed to
%alleviate this
%difficulty~\cite{martens11,pascanu2012difficulty,jaeger2004harnessing}
%with the most successful approach to date being Long Short Term Memory
%(LSTM)~\cite{hochreiter97}. 
%While LSTM RNNs are typically better than
%traditional RNNs, they are still difficult and unstable to train for
%dependencies beyond a few hundred timesteps with random
%initialization. 
For that reason, RNNs have rarely been used for natural language
processing tasks such as text classification despite their powers in
representing sequential structures.

On a number of document classification tasks, we find that it is
possible to train Long Short-Term Memory recurrent networks (LSTM
RNNs)~\cite{hochreiter97} to achieve good performance with careful
tuning of hyperparameters. Furthermore, a simple pretraining step can
significantly stabilize the training of LSTMs. For example, we can use
a next step prediction model, i.e., a {\em recurrent language model}
in NLP, as an unsupervised method. Another method is to use a {\em
  sequence autoencoder}, which uses a RNN to read a long input
sequence into a single vector. This vector will then be used to
reconstruct the original sequence. The weights obtained from these two
pretraining methods can then be used as an initialization for standard
LSTM RNNs to improve training and generalization.

In our experiments on document classification with 20
Newsgroups~\cite{Lang95} and DBpedia~\cite{lehmann2014dbpedia}, and
sentiment analysis with IMDB~\cite{Maas11} and Rotten
Tomatoes~\cite{Pang+Lee:05a}, LSTMs pretrained by recurrent language
models or sequence autoencoders are usually better than LSTMs
initialized randomly. This pretraining helps LSTMs reach or surpass
previous baselines on these datasets without additional data.

Another important result from our experiments is that using more
unlabeled data from related tasks in the pretraining can improve the
generalization of a subsequent supervised model. For example, using
unlabeled data from Amazon reviews to pretrain the sequence
autoencoders can improve classification accuracy on Rotten Tomatoes
from 79.7\% to 83.3\%, an equivalence of adding substantially more
labeled data. This evidence supports the thesis that it is possible to
use unsupervised learning with more unlabeled data to improve
supervised learning.  With sequence autoencoders, and outside
unlabeled data, LSTMs are able to match or surpass previously reported
results.

We believe our semi-supervised approach (as also argued
by~\cite{Ando2005}) has some advantages over other unsupervised
sequence learning methods, e.g., Paragraph
Vectors~\cite{le2014distributed}, because it can allow for easy
fine-tuning. Our semi-supervised learning approach is related to
Skip-Thought vectors~\cite{kiros2015skip}, with two differences. The
first difference is that Skip-Thought is a harder objective, because
it predicts adjacent sentences. The second is that Skip-Thought is a
pure unsupervised learning algorithm, without fine-tuning.

%It
%also suggests that a good principle for unsupervised learning is to
%use time as also argued by~\cite{hawkins2007intelligence}.

\section{Sequence autoencoders and recurrent language models}
\label{sec:model}
Our approach to sequence autoencoding is inspired by the work in
sequence to sequence learning (also known as {\em seq2seq}) by
Sutskever et al.~\cite{sutskever14}, which has been successfully used
for machine translation~\cite{luong2014addressing,jean2014using}, text
parsing~\cite{vinyals2014grammar}, image
captioning~\cite{vinyals2014show}, video
analysis~\cite{DBLP:journals/corr/SrivastavaMS15}, speech
recognition~\cite{chorowski2014end,chan2015listen} and conversational
modeling~\cite{shang2015neural,VinyalsL15}. Key to their approach is
the use of a recurrent network as an encoder to read in an input
sequence into a hidden state, which is the input to a decoder
recurrent network that predicts the output sequence.

\begin{figure}[h!]
\centering
\includegraphics[width=0.75\textwidth]{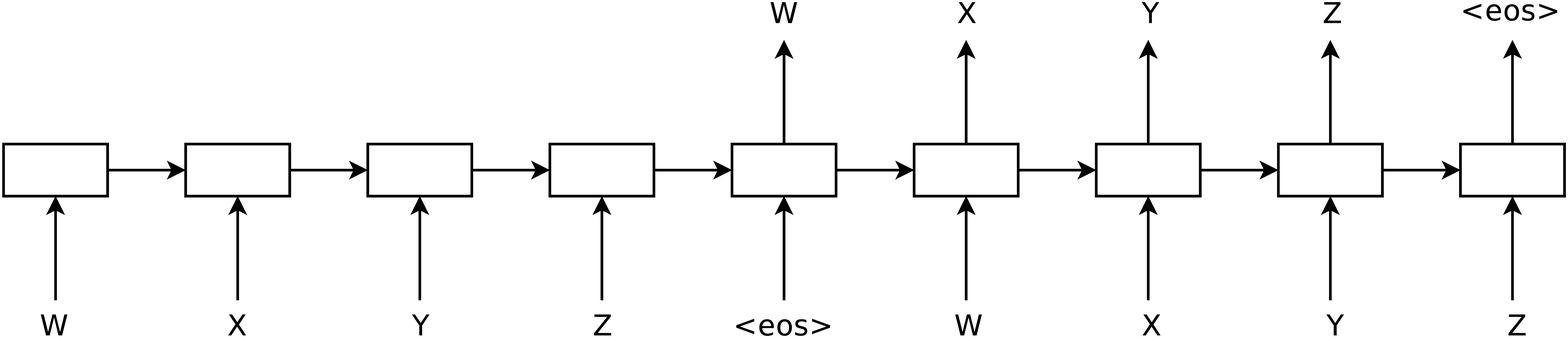}
\caption{The sequence autoencoder for the sequence ``WXYZ''. The
  sequence autoencoder uses a recurrent network to read the input
  sequence in to the hidden state, which can then be used to reconstruct the
  original sequence. }
\label{fig:sa}
\end{figure}

The sequence autoencoder is similar to the above concept, except that
it is an unsupervised learning model. The objective is to reconstruct
the input sequence itself. That means we replace the output sequence
in the {\em seq2seq} framework with the input sequence. In our
sequence autoencoders, the weights for the decoder network and the
encoder network are the same (see Figure~\ref{fig:sa}).

We find that the weights obtained from the sequence autoencoder can be
used as an initialization of another supervised network, one which
tries to classify the sequence. We hypothesize that this is because
the network can already memorize the input sequence. This reason, and
the fact that the gradients have shortcuts, are our hypothesis of why
the sequence autoencoder is a good and stable approach in initializing
recurrent networks.

A significant property of the sequence autoencoder is that it is
unsupervised, and thus can be trained with large quantities of
unlabeled data to improve its quality. Our result is that additional
unlabeled data can improve the generalization ability of recurrent
networks. This is especially useful for tasks that have limited
labeled data.

We also find that recurrent language
models~\cite{mikolov2010recurrent} can be used as a pretraining method
for LSTMs. This is equivalent to removing the encoder part of the
sequence autoencoder in Figure~\ref{fig:sa}. Our experimental results
show that this approach works better than LSTMs with random
initialization.

%Recurrent language models however sometimes perform
%slightly worse than sequence autoencoders, and hence we recommend to
%use sequence autoencoders as the default unsupervised pretraining
%method.

%We also find that LSTMs pretrained
%with a language model is much slower to converge than with a sequence
%autoencoder. Thus in the experiments, we report results of two
%methods, but mainly focus our attention to sequence autoencoders.

%\section{Overview of the datasets}
%Our primary goal is to improve the learning of long term dependencies
%in recurrent neural networks. For that purpose, we benchmarked
%recurrent networks on three tasks: document classification, sentiment
%analysis and machine translation. We will do so on publicly available
%and well understood datasets. The details are as follows.

\section{Overview of methods}
\label{sec:baselines}
In our experiments, we use LSTM recurrent networks~\cite{hochreiter97}
because they are generally better than RNNs. Our LSTM implementation
is standard and has input, forget, and output
gates~\cite{gers2000learning,DBLP:journals/corr/GreffSKSS15}. We
compare basic LSTMs against LSTMs initialized with the sequence
autoencoder method. When LSTMs are initialized with a sequence
autoencoder, the methods are called SA-LSTMs in our experiments. When
LSTMs are initialized with a language model, the method is called
LM-LSTMs.

%\begin{figure}[h!]
%\centering
%\includegraphics[width=0.34\textwidth]{lm}
%\caption{A neural language model for the sequence ``WXYZ''.}
%\label{fig:lm}
%\end{figure}

In most of our experiments our output layer predicts the document
label from the LSTM output at the last timestep. We also experiment
with the approach of putting the label at every timestep and linearly
increasing the weights of the prediction objectives from 0 to
1~\cite{DBLP:journals/corr/NgHVVMT15}. This way we can inject
gradients to earlier steps in the recurrent networks. We call this
approach {\em linear label gain}. Lastly, we also experiment with the
method of jointly training the supervised learning task with the
sequence autoencoder and call this method {\em joint training}.

\section{Experiments}
In our experiments with LSTMs, we follow the basic recipes as
described in~\cite{sutskever14} by clipping the cell outputs
and gradients. The benchmarks of focus are text understanding tasks,
with all datasets being publicly available. The tasks are sentiment
analysis (IMDB and Rotten Tomatoes) and text classification (20
Newsgroups and DBpedia). Commonly used methods on these datasets, such
as bag-of-words or n-grams, typically ignore long-range ordering
information (e.g., modifiers and their objects may be separated by
many unrelated words); so one would expect recurrent methods which
preserve ordering information to perform well. Nevertheless, due to
the difficulty in optimizing these networks, recurrent models are not
the method of choice for document classification.

In our experiments with the sequence autoencoder, we train 
it to reproduce the full document after reading all
the input words. In other words, we do not perform any truncation or
windowing. We add an end of sentence marker to the end of each input
sequence and train the network to start reproducing the sequence after
that marker. To speed up performance and reduce GPU memory usage, we
perform truncated backpropagation up to 400 timesteps from the end of
the sequence. We preprocess the text so that punctuation is treated
as separate tokens and we ignore any non-English characters and words
in the DBpedia text. We also remove words that only appear once in
each dataset and do not perform any term weighting or stemming.

After training the recurrent language model or the sequence autoencoder for
roughly 500K steps with a batch size of 128, we use both the
word embedding parameters and the LSTM weights to initialize
the LSTM for the supervised task. We then train on that task
while fine tuning both the embedding parameters and the weights and
use early stopping when the validation error starts to increase. We
choose the dropout parameters based on a validation set.

Using SA-LSTMs, we are able to match or surpass reported results for
all datasets. It is important to emphasize that previous best results
come from various different methods. So it is significant that one
method achieves strong results for all datasets, presumably because
such a method can be used as a general model for any similar
task. A summary of results in the experiments are shown in
Table~\ref{table:summary}. More details of the experiments are as
follows.
\begin{table}[h!]
\centering
\caption{A summary of the error rates of SA-LSTMs and previous best
  reported results.}
\label{table:summary}
\begin{tabular}{lrr}
\toprule
\multicolumn{1}{l}{\bf Dataset}           &  \multicolumn{1}{r}{\bf SA-LSTM } & \multicolumn{1}{r}{\bf Previous best result } \\
\midrule
IMDB                    &  7.24\%           & 7.42\%                         \\
Rotten Tomatoes         &  16.7\%           & 18.5\%                         \\
20 Newsgroups           &  15.6\%           & 17.1\%                          \\
DBpedia                 &  1.19\%           & 1.74\%                          \\
\bottomrule
\end{tabular}
\end{table}

\subsection{Sentiment analysis experiments with IMDB}
In this first set of experiments, we benchmark our methods on the IMDB
movie sentiment dataset, proposed by Maas et
al.~\cite{Maas11}.\footnote{\url{http://ai.Stanford.edu/amaas/data/sentiment/index.html}}
There are 25,000 labeled and 50,000 unlabeled documents in the
training set and 25,000 in the test set. We use 15\% of the labeled
training documents as a validation set. The average length of each
document is 241 words and the maximum length of a document is 2,526
words. The previous baselines are bag-of-words,
ConvNets~\cite{kim2014convolutional} or Paragraph
Vectors~\cite{le2014distributed}.

Since the documents are long, one might expect that it is difficult
for recurrent networks to learn. We however find that with
tuning, it is possible to train LSTM recurrent networks to fit the
training set. For example, if we set the size of hidden state to be
512 units and truncate the backprop to be 400, an LSTM can do fairly
well. With random embedding dimension
dropout~\cite{zaremba2014recurrent} and random word dropout (not
published previously), we are able to reach performance of around
86.5\% accuracy in the test set, which is approximately 5\% worse than most
baselines.

Fundamentally, the main problem with this approach is that it is
unstable: if we were to increase the number of hidden units or to
increase the number of backprop steps, the training breaks down very
quickly: the objective function explodes even with careful tuning of
the gradient clipping. This is because LSTMs are sensitive to the
hyperparameters for long documents. In contrast, we find that the
SA-LSTM works better and is more stable. If we use the sequence
autoencoders, changing the size of the hidden state or the number of
backprop steps hardly affects the training of LSTMs. This is important
because the models become more practical to train.

\iffalse
As sequence autoencoders work so well, one might wonder if there must
be something wrong with the initialization of the LSTM in our
implementation and whether it is possible to improve that. We hence
ran a control experiment where we took the weights from an SA-LSTM,
computed the variance of the weights and then used a random Gaussian
initialization scaled by the computed variance. If this initialization
works well, one may want to change the scaling factors in the random
initialization and use that in lieu of the SA-LSTM. The result is that
the new scaling factors help training, but not to the extent that
they can replace sequence autoencoders. The results are shown in
Figure~\ref{fig:imdb}.
\begin{figure}[h!]
\centering
\includegraphics[width=0.4\textwidth]{train_imdb}
%%\includegraphics[width=0.45\textwidth]{test_imdb}
\caption{Training error rates during optimization of LSTMs, LM-LSTMs
  and SA-LSTMs. Without the sequence autoencoder initialization at the
  beginning, LSTMs fail to converge. LM-LSTMs, SA-LSTMs converge much
  better than LSTMs. SA-LSTMs converge slightly faster than LM-LSTMs
  because they have better long term effects. LSTM-scaled is the
  method that uses scaling factors computed from the weights of the
  trained sequence autoencoder.}
\label{fig:imdb}
\end{figure}
\fi

Using sequence autoencoders, we overcome the optimization instability
in LSTMs in such a way that it is fast and easy to achieve perfect
classification on the training set. To avoid overfitting, we again use
input dimension dropout, with the dropout rate chosen on a validation
set. We find that dropping out 80\% of the input embedding dimensions
works well for this dataset. The results of our experiments are shown
in Table~\ref{imdb-table} together with previous baselines. We also add an
additional baseline where we initialize a LSTM with word2vec embeddings on the
training set.

\begin{table}[h!]
  \caption{Performance of models on the IMDB sentiment classification task.}
  \label{imdb-table}
  \begin{center}
    \begin{tabular}{ld{2}}
      \toprule
      \multicolumn{1}{c}{\bf Model} & \multicolumn{1}{c}{\bf Test error rate}\\
      \midrule
      LSTM with tuning and dropout   & 13.50\%\\
      LSTM initialized with word2vec embeddings  & 10.00\%\\
      %%sum-LSTM (see Figure~\ref{fig:clstm}) & 50.00\%\\
      LM-LSTM (see Section~\ref{sec:model}) & 7.64\%\\
      SA-LSTM (see Figure~\ref{fig:sa}) & 7.24\% \\
      SA-LSTM with linear gain (see Section~\ref{sec:baselines}) & 9.17\% \\
      SA-LSTM with joint training (see Section~\ref{sec:baselines}) &  14.70\% \\
      \midrule
      Full+Unlabeled+BoW~\cite{Maas11}& 11.11\% \\
      WRRBM + BoW (bnc)~\cite{Maas11} & 10.77\% \\
      NBSVM-bi (Na\"ive Bayes SVM with bigrams)~\cite{sidaw12simple} & 8.78\% \\
      seq2-bow{\em n}-CNN (ConvNet with dynamic pooling)~\cite{johnson2014effective} & 7.67\% \\
      Paragraph Vectors~\cite{le2014distributed} & 7.42\% \\
      \bottomrule
      \end{tabular}
    \end{center}
  \end{table}

The results confirm that SA-LSTM with input embedding dropout can be
as good as previous best results on this dataset. In contrast, LSTMs
without sequence autoencoders have trouble in optimizing the objective
because of long range dependencies in the documents.

Using language modeling (LM-LSTM) as an initialization works well,
achieving 8.98\%, but less well compared to the SA-LSTM. This is
perhaps because language modeling is a short-term objective, so that
the hidden state only captures the ability to predict the next few
words. 

In the above table, we use 1,024 units for memory cells, 512 units for
the input embedding layer in the LM-LSTM and SA-LSTM.  We also use a
hidden layer 512 units with dropout of 50\% between the last hidden
state and the classifier. We continue to use these settings in the
following experiments except with 30 units in the final hidden layer.

In Table~\ref{imdb-examples-table}, we present some examples from the IMDB dataset that are correctly classified by SA-LSTM but not by a bigram NBSVM model. These examples often have long-term dependencies or have sarcasm that is difficult to find by solely looking at short phrases.

\begin{table}[h!]
  \begin{small}

  \caption{IMDB sentiment classification examples that are correctly classified by SA-LSTM and incorrectly by NBSVM-bi.}
  \label{imdb-examples-table}
  \begin{center}
    \begin{tabular}{p{11cm}l}
      \toprule
      \multicolumn{1}{c}{\bf Text} & \multicolumn{1}{c}{\bf Sentiment}\\
      \midrule
      This film is not at all as bad as some people on here are saying. I think it has got a decent horror plot and the acting seem normal to me. People are way over-exagerating what was wrong with this. It is simply classic horror, the type without a plot that we have to think about forever and forever. We can just sit back, relax, and be scared. & Positive \\
      \midrule
       Looking for a REAL super bad movie? If you wanna have great fun, don't hesitate and check this one! Ferrigno is incredibly bad but is also the best of this mediocrity. & Negative\\
      \midrule
      A professional production with quality actors that simply never touched the heart or the funny bone no matter how hard it tried. The quality cast, stark setting and excellent cinemetography made you hope for Fargo or High Plains Drifter but sorry, the soup had no seasoning...or meat for that matter. A 3 (of 10) for effort. & Negative\\
      \midrule
      The screen-play is very bad, but there are some action sequences that i really liked. I think the image is good, better than other romanian movies. I liked also how the actors did their jobs. & Negative\\
      \bottomrule
      \end{tabular}
    \end{center}
    \end{small}
  \end{table}

\subsection{Sentiment analysis experiments with Rotten Tomatoes and the positive effects of additional unlabeled data}
The success on the IMDB dataset convinces us to test our methods on
another sentiment analysis task to see if similar gains can be
obtained. The benchmark of focus in this experiment is the Rotten Tomatoes
dataset~\cite{Pang+Lee:05a}.\footnote{\url{http://www.cs.cornell.edu/people/pabo/movie-review-data/}}
The dataset has 10,662 documents, which are randomly split into 80\%
for training, 10\% for validation and 10\% for test. The average
length of each document is 22 words and the maximum length is 52
words. Thus compared to IMDB, this dataset is smaller both in terms of
the number of documents and the number of words per document.

Our first observation is that it is easier to train LSTMs on this
dataset than on the IMDB dataset and the gaps between LSTMs, LM-LSTMs
and SA-LSTMs are smaller than before. This is because movie reviews in
Rotten Tomatoes are sentences whereas reviews in IMDB are paragraphs.

%However, SA-LSTMs are still more stable than LSTMs.

As this dataset is small, our methods tend to severely overfit the
training set. Combining SA-LSTMs with 95\% input embedding and 50\%
word dropout improves generalization and allows the model to achieve
20.3\% test set error. %, which is best amongst methods that do not use
%outside data.
%A carefully-tuned LSTM only reached 21.0\%.
Further tuning the hyperparameters on the validation set yields 19.3\% test error.

To better the performance, we add unlabeled data from the IMDB dataset
in the previous experiment and Amazon movie
reviews~\cite{mcauley2013hidden} to the autoencoder training
stage.\footnote{The dataset is available at
  \url{http://snap.stanford.edu/data/web-Amazon.html}, which has 34
  million general product reviews, but we only use 7.9 million movie
  reviews in our experiments.}  We also run a control experiment where we
use the pretrained word vectors trained by word2vec from Google News.
\begin{table}[h!]
  \caption{Performance of models on the Rotten Tomatoes sentiment classification task.}
  \label{rt-table}
  \begin{center}
    \begin{tabular}{ld{1}}
      \toprule
      \multicolumn{1}{c}{\bf Model} & \multicolumn{1}{c}{\bf Test error rate}\\
      \midrule
      LSTM with tuning and dropout & 20.3\% \\
      LSTM with linear gain & 21.1\%\\
      LM-LSTM & 21.7\% \\
      SA-LSTM & 20.3\% \\
      \midrule
      LSTM with word vectors from word2vec Google News & 20.5\% \\
      SA-LSTM with unlabeled data from IMDB & 18.6\%\\
      SA-LSTM with unlabeled data from Amazon reviews & 16.7\%\\
      \midrule
      MV-RNN~\cite{socher2012semantic} & 21.0\% \\
      NBSVM-bi~\cite{sidaw12simple} & 20.6\% \\
      CNN-rand~\cite{kim2014convolutional} & 23.5\% \\
      CNN-non-static (ConvNet with vectors from word2vec Google News)~\cite{kim2014convolutional} & 18.5\% \\
      \bottomrule
      \end{tabular}
    \end{center}
  \end{table}

The results for this set of experiments are shown in
Table~\ref{rt-table}. Our observation is that if we use the word
vectors from word2vec, there is only a small gain of 0.5\%. This is
perhaps because the recurrent weights play an important role in our
model and are not initialized properly in this experiment. However, if
we use IMDB to pretrain the sequence autoencoders, the error decreases
from 20.5\% to 18.6\%, nearly a 2\% gain in accuracy; if we use Amazon
reviews, a larger unlabeled dataset (7.9 million movie reviews), to
pretrain the sequence autoencoders, the error goes down to 16.7\%
which is another 2\% gain in accuracy.

This brings us to the question of how well this method of using
unlabeled data fares compared to adding more labeled data. As argued
by Socher et al.~\cite{socher13}, a reason of why the methods are not
perfect yet is the lack of labeled training data, they proposed to use
more labeled data by labeling an addition of 215,154 phrases created
by the Stanford Parser. The use of more labeled data allowed their
method to achieve around 15\% error in the test set, an improvement of
approximately 5\% over older methods with less labeled data.

We compare our method to their reported results~\cite{socher13} on
sentence-level classification. As our method does not have access to
valuable labeled data, one might expect that our method is severely
disadvantaged and should not perform on the same level. However, with
unlabeled data and sequence autoencoders, we are able to obtain
16.7\%, ranking second amongst many other methods that have access to
a much larger corpus of labeled data. The fact that unlabeled data can
compensate for the lack of labeled data is very significant as
unlabeled data are much cheaper than labeled data. The results are
shown in Table~\ref{rt-table-unlabeled}.

\begin{table}[h!]
  \caption{More unlabeled data vs. more labeled data. Performance of
    SA-LSTM with additional unlabeled data and previous models with
    additional labeled data on the Rotten Tomatoes task.}
  \label{rt-table-unlabeled}
  \begin{center}
    \begin{tabular}{ld{1}}
      \toprule
      \multicolumn{1}{c}{\bf Model} & \multicolumn{1}{c}{\bf Test error rate}\\
      \midrule
      LSTM initialized with word2vec embeddings trained on Amazon reviews & 23.3\%\\
      SA-LSTM with unlabeled data from Amazon reviews & 16.7\%\\
      \midrule
      NB~\cite{socher13} & 18.2\%\\
      SVM~\cite{socher13} & 20.6\%\\
      BiNB~\cite{socher13} & 16.9\%\\
      VecAvg~\cite{socher13} & 19.9\%\\
      RNN~\cite{socher13} & 17.6\%\\
      MV-RNN~\cite{socher13} & 17.1\%\\
      RNTN~\cite{socher13} & 14.6\%\\
      \bottomrule
      \end{tabular}
    \end{center}
  \end{table}

\subsection{Text classification experiments with 20 newsgroups}
The experiments so far have been done on datasets where the number of
tokens in a document is relatively small, a few hundred words. Our
question becomes whether it is possible to use SA-LSTMs for tasks that
have a substantial number of words. For that purpose, we carry out the
next experiments on the 20 newsgroups
dataset~\cite{Lang95}.\footnote{\url{http://qwone.com/~jason/20Newsgroups/}}
There are 11,293 documents in the training set and 7,528 in the test
set. We use 15\% of the training documents as a validation set. Each
document is an email with an average length of 267 words and a maximum
length of 11,925 words.  Attachments, PGP keys, duplicates and empty
messages are removed.  As the newsgroup documents are long, it was
previously considered improbable for recurrent networks to learn
anything from the dataset. The best methods are often simple
bag-of-words.

We repeat the same experiments with LSTMs and SA-LSTMs on this
dataset. Similar to observations made in previous experiments,
SA-LSTMs are generally more stable to train than LSTMs. To improve
generalization of the models, we again use input embedding dropout
and word dropout chosen on the validation set. With 70\% input embedding dropout and 75\% word
dropout, SA-LSTM achieves 15.6\% test set error which is much better
than previous classifiers in this dataset. Results are shown in
Table~\ref{20ng-table}.

\begin{table}[h!]
  \caption{Performance of models on the 20 newsgroups classification task.}
  \label{20ng-table}
  \begin{center}
    \begin{tabular}{ld{2}}
      \toprule
      \multicolumn{1}{c}{\bf Model} & \multicolumn{1}{c}{\bf Test error rate}\\
      \midrule
      LSTM & 18.0\% \\
      LSTM with linear gain & 71.6\%\\
      LM-LSTM & 15.3\% \\
      SA-LSTM & 15.6\% \\
      \midrule
      Hybrid Class RBM~\cite{larochelle2012learning}& 23.8\% \\
      RBM-MLP~\cite{dauphin2013stochastic} & 20.5\% \\
      SVM + Bag-of-words~\cite{cardoso} & 17.1\%\\
      Na\"ive Bayes~\cite{cardoso} & 19.0\%\\
      \bottomrule
      \end{tabular}
    \end{center}
  \end{table}

\subsection{Character-level document classification experiments with DBpedia}

In this set of experiments, we turn our attention to another
challenging task of categorizing Wikipedia pages by reading
character-by-character inputs. The dataset of attention is the DBpedia
dataset~\cite{lehmann2014dbpedia}, which was also used to benchmark
convolutional neural nets in Zhang and
LeCun~\cite{DBLP:journals/corr/ZhangL15}.
%Note that unlike other datasets in Zhang and LeCun~\cite{DBLP:journals/corr/ZhangL15},
DBpedia had no duplication or tainting issues from the outset so we compare
experimental results on this dataset. DBpedia is a
crowd-sourced effort to extract information from Wikipedia and
categorize it into an ontology.

For this experiment, we follow the same procedure suggested in Zhang
and LeCun~\cite{DBLP:journals/corr/ZhangL15}. The task is to classify
DBpedia abstracts into one of 14 categories after reading the
character-by-character input. The dataset is split into 560,000
training examples and 70,000 test examples. A DBpedia document has an
average of 300 characters while the maximum length of all documents is
13,467 characters. As this dataset is large, overfitting is not an
issue and thus we do not perform any dropout on the input or
recurrent layers. For this dataset, we use a two-layered LSTM, each
layer has 512 hidden units and and the input embedding has 128 units.

\begin{table}[h!]
  \caption{Performance of models on the DBpedia character level classification task.}
  \label{dbpedia-table}
  \begin{center}
    \begin{tabular}{lcd{2}}
      \toprule
      \multicolumn{1}{c}{\bf Model} & \multicolumn{1}{c}{\bf Test error rate}\\
      \midrule
      LSTM & 13.64\% \\
      LSTM with linear gain & 1.32\% \\
      LM-LSTM & 1.50\% \\
      SA-LSTM & 2.34\% \\
      SA-LSTM with linear gain & 1.23\% \\
      SA-LSTM with 3 layers and linear gain & 1.19\% \\
      SA-LSTM (word-level) & 1.40\% \\
      \midrule
      Bag-of-words & 3.57\% \\
      Small ConvNet & 1.98\% \\
      Large ConvNet & 1.73\% \\
      \bottomrule
      \end{tabular}
    \end{center}
  \end{table}

In this dataset, we find that the linear label gain as described in
Section~\ref{sec:baselines} is an effective mechanism to inject
gradients to earlier steps in LSTMs. This linear gain method works
well and achieves 1.32\% test set error, which is better than
SA-LSTM. Combining SA-LSTM and the linear gain method achieves 1.19\%
test set error, a significant improvement from the results of
convolutional networks as shown in Table~\ref{dbpedia-table}.

\subsection{Object classification experiments with CIFAR-10}

In these experiments, we attempt to see if our pre-training methods
extend to non-textual data. To do this, we train a LSTM on the
CIFAR-10 image dataset, consisting of 60,000 32x32 colour images divided into
10 classes.
The input at each timestep of the LSTM is
an entire row of pixels and we predict the class of the image after reading the
final row. We use the same method as in~\cite{kriz12} to perform data
augmentation.  We also trained a LSTM to do next row prediction given
the current row (we denote this as LM-LSTM) and a LSTM to autoencode the
image by rows (SA-LSTM). The loss function
during unsupervised learning is the Euclidean L2 distance between the
predicted and the target row. We then fine-tune these on the
classification task and present the classification results in
Table~\ref{cifar-table}. While we do not achieve the results attained
by state of the art convolutional networks, our 2-layer pretrained
LM-LSTM is able to exceed the results of the baseline convolutional
DBN model~\cite{krizhevskycifar10} despite not using any convolutions
and outperforms the non pre-trained LSTM.

\begin{table}[h!]
  \caption{Performance on the CIFAR-10 object classification task with 50,000 train images.}
  \label{cifar-table}
  \begin{center}
    \begin{tabular}{lcd{2}}
      \toprule
      \multicolumn{1}{c}{\bf Model (400 units)} & \multicolumn{1}{c}{\bf Test error rate}\\
      \midrule
      1-layer LSTM & 25.0\% \\
      1-layer LM-LSTM & 23.1\% \\ % 76.83
      1-layer SA-LSTM & 25.1\% \\
      \midrule
      2-layer LSTM & 26.0\% \\
      2-layer LM-LSTM & 18.0\% \\
      2-layer SA-LSTM & 25.1\% \\
      \midrule
      Convolution DBNs~\cite{krizhevskycifar10} & 21.1\% \\
      \bottomrule
      \end{tabular}
    \end{center}
  \end{table}

\section{Discussion}
In this paper, we showed that it is possible to use LSTM recurrent
networks for NLP tasks such as document classification. Further, we
demonstrated that a language model or a sequence autoencoder can help
stabilize the learning in LSTM recurrent networks. On five benchmarks
that we tried, LSTMs can reach or surpass the performance levels of
all previous baselines.

%The ability to learn long range dependencies perhaps will have more
%impact beyond the text understanding tasks in this paper. We believe
%that information retrieval and ranking, video analysis and time series
%analysis are direct applications of the method.

\paragraph{Acknowledgements:}
We thank Oriol Vinyals, Ilya Sutskever, Greg Corrado, Vijay Vasudevan,
Manjunath Kudlur, Rajat Monga, Matthieu Devin, and the Google Brain
team for their help.

\bibliography{translate}
\bibliographystyle{plain}
\end{document}